\documentclass[letterpaper, 10 pt, conference]{ieeeconf}  % Comment this line out if you need a4paper
\usepackage[cmex10]{amsmath}
\usepackage{times}
\usepackage{epsfig}
\usepackage{graphicx}
\usepackage{amssymb}
\usepackage{algorithmic}
\usepackage{algorithm}
\usepackage{color}
\usepackage{subfigure}
\usepackage{tikz}
\usepackage{mathrsfs}
\usepackage{bm}
\usepackage{multicol}
\usepackage{multirow}
\usepackage[section]{placeins}
\usepackage{url}
\usepackage{array}

\definecolor{fgreen}{RGB}{34,139,34}

\IEEEoverridecommandlockouts                              % This command is only needed if 
\title{\LARGE \bf
Heart Beat Characterization from Ballistocardiogram Signals using Extended Functions of Multiple Instances
}

\author{Changzhe Jiao,~\IEEEmembership{Student Member,~IEEE,} Princess Lyons,~\IEEEmembership{Student Member,~IEEE,} Alina Zare,\\~\IEEEmembership{Senior Member,~IEEE},  Licet Rosales,~\IEEEmembership{Student Member,~IEEE,} and Marjorie Skubic,~\IEEEmembership{Senior Member,~IEEE}% <-this % stops a space
\thanks{This material is based upon work supported by the National Science Foundation under Grant No. IIS-1350078 - CAREER: Supervised Learning for Incomplete and Uncertain Data. The study was approved by the University of Missouri Institutional Review Board.}% <-this % stops a space
\thanks{Changzhe Jiao, Princess Lyons, Alina Zare, Licet Rosales and Marjorie Skubic are with the ECE Dept., University of Missouri
(e-mail: cjr25@mail.missouri.edu, pplprd@mail.missouri.edu, zarea@missouri.edu, lr5zf@mail.missouri.edu and SkubicM@missouri.edu).}%
}

\begin{document}

\maketitle
\thispagestyle{empty}
\pagestyle{empty}

%%%%%%%%%%%%%%%%%%%%%%%%%%%%%%%%%%%%%%%%%%%%%%%%%%%%%%%%%%%%%%%%%%%%%%%%%%%%%%%%
\begin{abstract}

A multiple instance learning (MIL) method, extended Function of Multiple Instances ($e$FUMI), is applied to ballistocardiogram (BCG) signals produced by a hydraulic bed sensor. The goal of this approach is to learn a personalized heartbeat ``concept'' for an individual.  This heartbeat concept is a prototype (or ``signature'') that characterizes the heartbeat pattern for an individual in ballistocardiogram data.  The $e$FUMI method models the problem of learning a heartbeat concept from a BCG signal as a MIL problem. This approach elegantly addresses the uncertainty inherent in a BCG signal (\textit{e. g.}, misalignment between training data and ground truth, mis-collection of heartbeat by some transducers, \textit{etc.}). Given a BCG training signal coupled with a ground truth signal (e.g., a pulse finger sensor), training ``bags'' labeled with only binary labels denoting if a training bag contains a heartbeat signal or not can be generated.  Then, using these bags, $e$FUMI learns a personalized concept of heartbeat for a subject as well as several non-heartbeat background concepts. After learning the heartbeat concept, heartbeat detection and heart rate estimation can be applied to test data. Experimental results show that the estimated heartbeat concept found by $e$FUMI is more representative and a more discriminative prototype of the heartbeat signals than those found by comparison MIL methods in the literature.
\end{abstract}

%%%%%%%%%%%%%%%%%%%%%%%%%%%%%%%%%%%%%%%%%%%%%%%%%%%%%%%%%%%%%%%%%%%%%%%%%%%%%%%%
\vspace{-1mm}
\section{INTRODUCTION}

Long-term measurement and monitoring of vital signs, \textit{e. g.} heart rate, respiratory rate, body temperature and blood pressure, \textit{etc.}, provides promise for the early treatment of any potential problems, especially for older adults. Compared with the many wearable heartrate monitoring systems available, ballistocardiography provides an unintrusive and, thus, comfortable monitoring alternative.  These systems record the motion of the human body generated by the sudden ejection of blood into the large vessels at each cardiac cycle \cite{starr1939studies}. This technique has gained revived interest due to recent development in measurement technology \cite{pinheiro2010theory, inan2015ballistocardiography} and a growing preference for ``aging in place'' \cite{rantz2005technology}.

A hydraulic bed sensor system \cite{rosales2012heartbeat, heise2013non} has been recently developed to non-invasively collect the superposition of the ballistocardiogram (BCG) and respiration signals. The motivation for this work is to support continuous, non-intrusive monitoring of vital signs of older adults in an unstructured natural living environment. The hydraulic bed sensor is placed beneath the mattress and has shown to be flexible, low-cost and non-intrusive for monitoring an individuals heartrate. Compared with other methods such as electrocardiography (ECG), BCG does not need electrodes or clips to be affixed to the patient's body and thus is ideal for long term in-home monitoring. However, the lack of saliency and large variability in a BCG signal makes it much more difficult to detect individual heartbeats than with an ECG. 

Multiple-instance learning (MIL) is a variation on supervised learning for problems with incomplete training label information \cite{Dietterich:1997, Maron:1998, Bolton:2011}. In MIL, training data are grouped into positive and negative ``bags.'' A bag is defined to be a multi-set of data points in which a positive bag contains at least one data point from the target class and negative bags are composed entirely of non-target data. Thus, data point-specific training labels are unavailable.  Given training data in this form, the majority of MIL methods  either: (1) learn target concepts for describing the target class; or (2) train a classifier that can distinguish between individual target and non-target data points. Here, \emph{concepts} refer to generalized class prototypes in the feature space; in the case of heartbeat characterization, a concept is the discriminative heartbeat pattern of an individual.

This paper proposes to estimate a personalized heartbeat concept and, given this concept, perform heartbeat detection and heartrate estimation from BCG signals.  The proposed method for addressing this goal is the application of the extended Function of Multiple Instances ($e$FUMI) algorithm \cite{Zare:2014whispers, Zare:2015fumi}. $e$FUMI is well suited to the problem of heartbeat characterization from BCG signals as it is able to address multiple types of uncertainty in training data.  Namely, $e$FUMI addresses mixed data samples and inaccurate labels in training data; both are common in BCG analysis. 

\vspace{-1mm}
\section{HYDRAULIC BED SENSOR}\label{sec:2_sensor}

\begin{figure}
\vspace{2mm}
\begin{center}
\subfigure[Sensor and Embedded System]{   
\includegraphics[width=3.3cm]{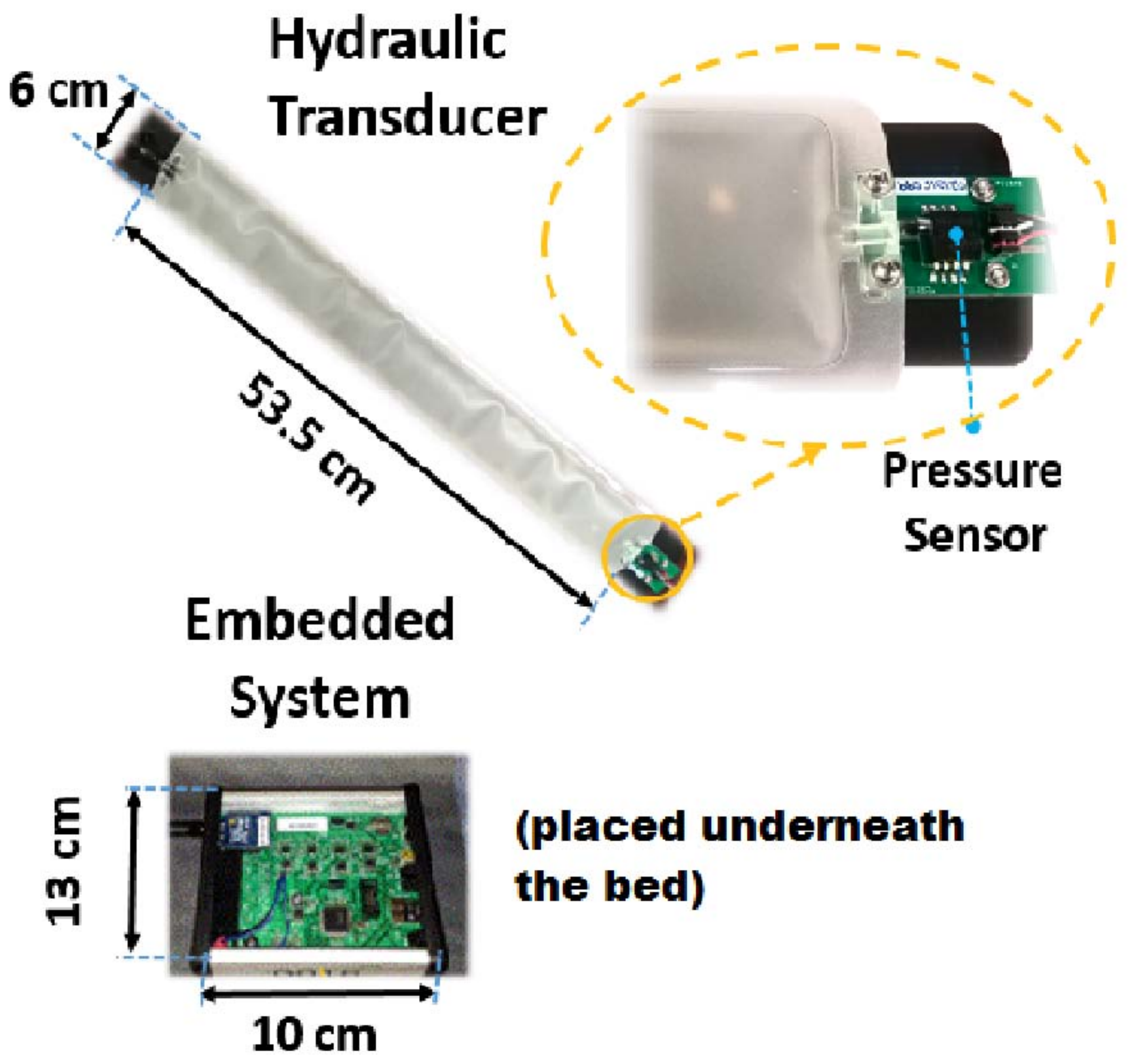} \label{fig:bed_system}}
\subfigure[Transducer placement]{   
\includegraphics[width=4.3cm]{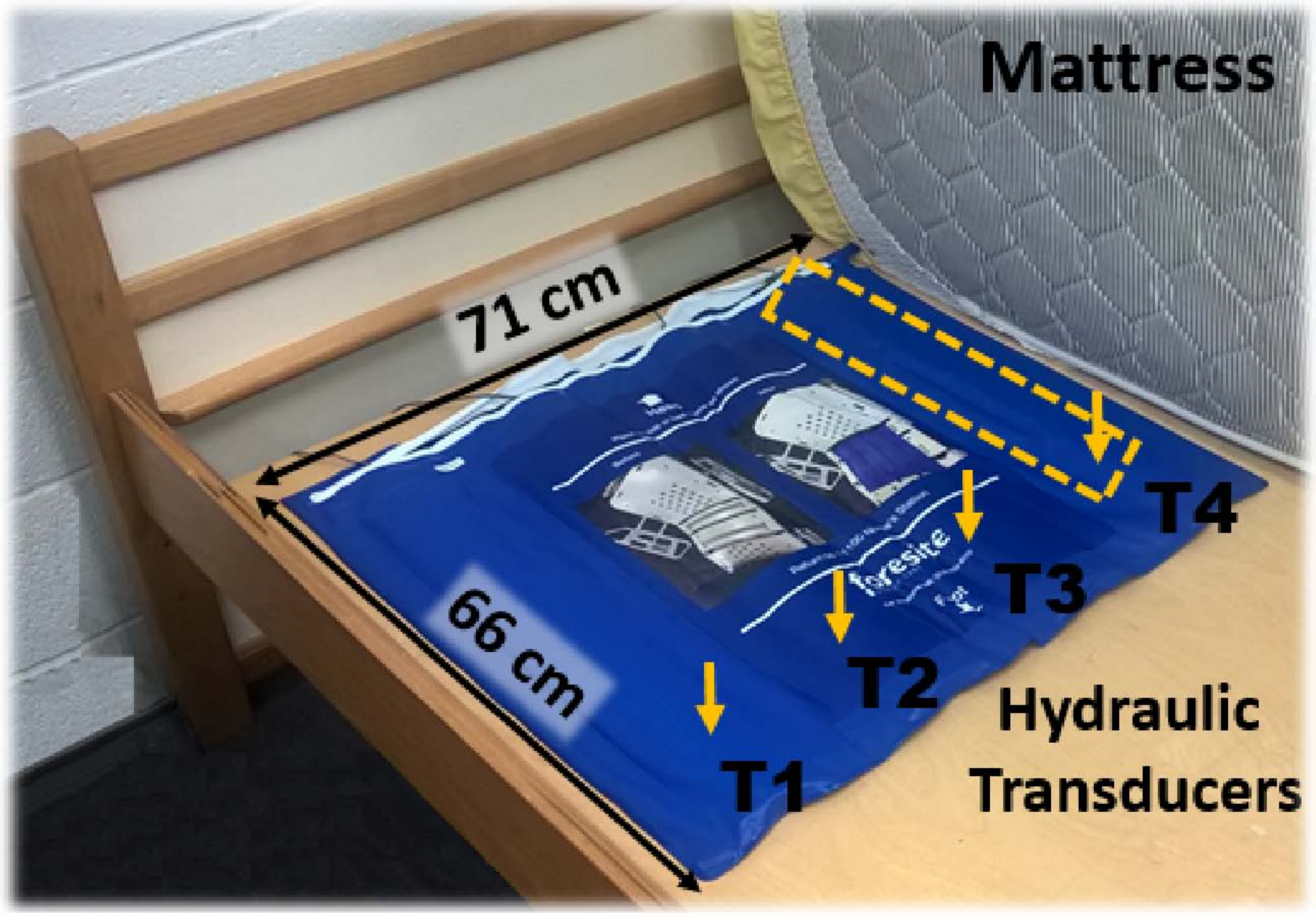} \label{fig:bed_transducer}}
\caption{Hydraulic Bed Sensor System}\label{fig:bed_sensor}
\end{center}
\end{figure}

The hydraulic bed sensor is composed of a transducer and a pressure sensor. The transducer is 53.5 cm long and 6 cm wide, and is filled with 0.4 liter of water \cite{rosales2012heartbeat, heise2013non}. The integrated silicon pressure sensor (Freescale MPX5010GP) attached to the end of the transducer is used for measuring the vibration of the discharged hose. Four transducers are placed in parallel underneath the mattress to provide enough coverage of the pulse from the heartbeat. Fig. \ref{fig:bed_system} and \ref{fig:bed_transducer}  show the hydraulic bed sensor system and placement of the 4 transducers, respectively. The four transducers are identical, but the data quality collected by those 4 transducers could vary depending on the sleeping position, type of mattress and the physical characteristics of the subject.

\begin{figure}
\begin{center}
\includegraphics[width=7cm]{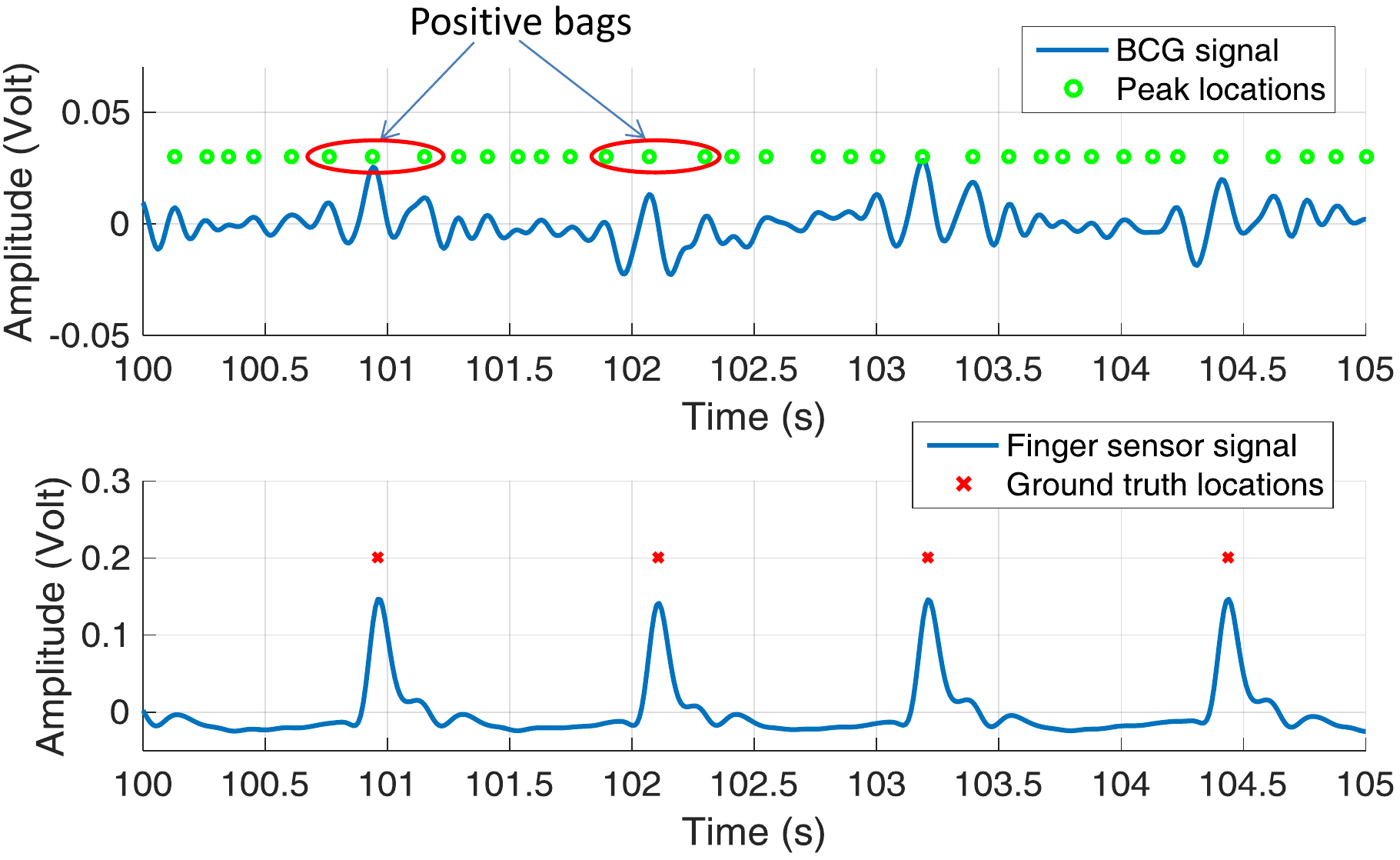}
\caption{BCG Signal and Ground Truth Plot}\label{fig:BCG_FS_plot}
\end{center}
\vspace{-5mm}
\end{figure}

\begin{figure}
\begin{center}
\includegraphics[width=7cm]{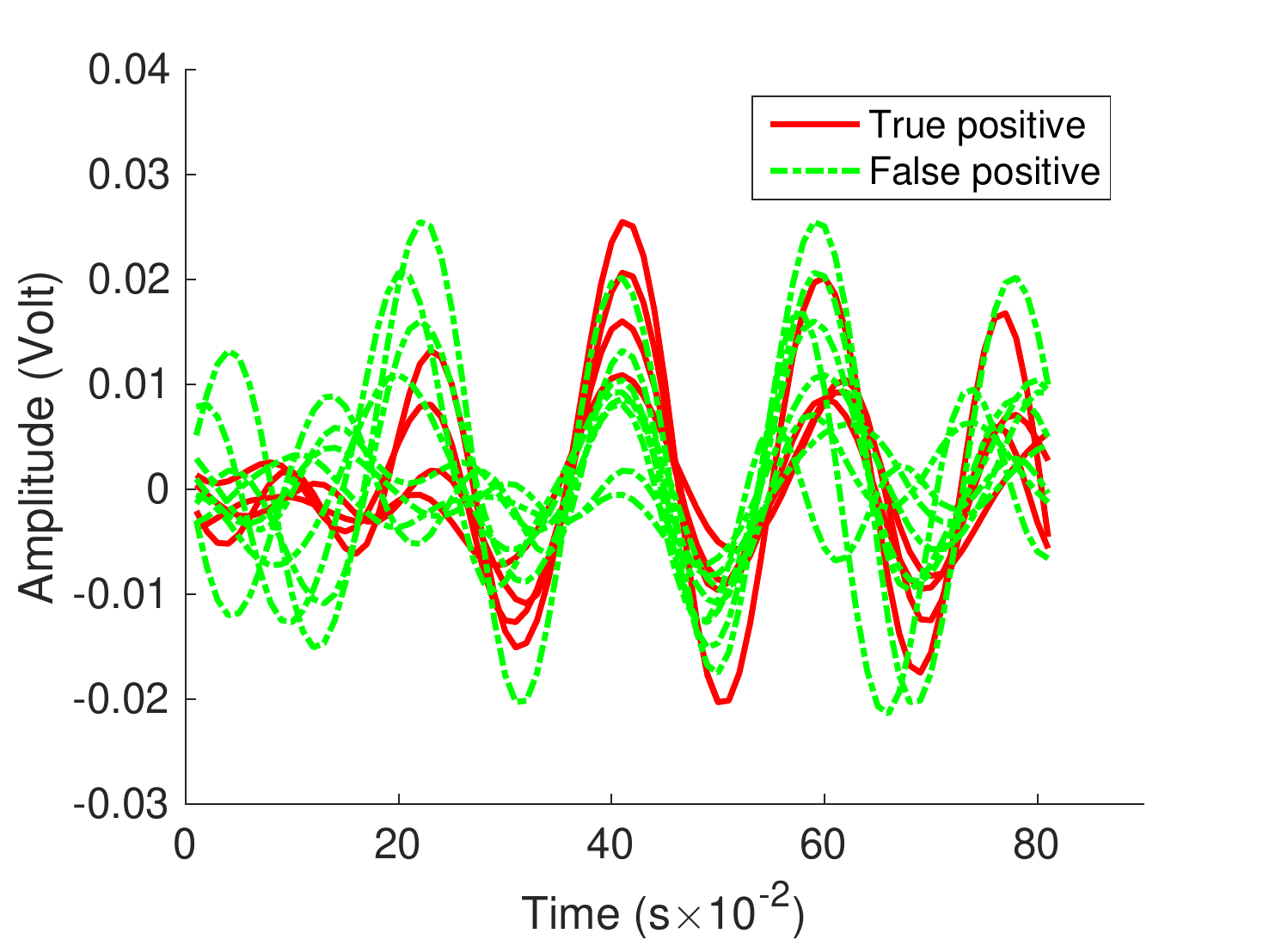}
\caption{Plot of One Positive Bag where the true positive heartbeat signals tend to have more prominent J-peak }\label{fig:pos_bag_plot}
\end{center}
\end{figure}

Given the described hydraulic bed sensor, a BCG data set was collected for several people lying on their backs in the bed setup described above for 10 minutes each and a subset data containing BCG signal of four subjects was tested by the proposed method. The BCG signal was sampled at 100 Hz and filtered by a Butterworth band-pass filter with 3dB cutoff at 0.4 Hz and 10 Hz  to remove respiratory effects and high frequency noise. The study was approved by the University of Missouri Institutional Review Board.

Also, for ground-truthing, a piezoelectric pulse sensor (ADInstruments MLT1010) attached to subject's finger is used to record the pulse ejected by a heartbeat.  Fig. \ref{fig:BCG_FS_plot} shows a typical filtered BCG signal collected by one transducer and the corresponding finger sensor ground truth information, where the green circles denote every peak location of the filtered BCG signal. In this figure, we can see that there are prominent peak patterns, that appear near the ground truth locations denoted by finger sensor. However, although both of the sensors are assumed to be capturing a corresponding heartbeat signal simultaneously, there is some misalignment between the finger sensor and BCG pressure sensor. Also, it is possible that one or more transducers fail to collect a clear heartbeat signal depending on the location and position of the subject in the bed. These uncertainties in the training data make it more difficult to apply traditional supervised learning methods for heartbeat detection or heart rate estimation on BCG signal. 

In this paper, we propose to introduce the idea of training ``bags'' to address label uncertainty as well as mis-collection of heartbeat signals in the BCG data. Specifically, we extract several BCG sub-segments (each 81 points long corresponding to 0.81s) that are centered on each peak in the BCG training data.  Each of these sub-segments are then treated as individual data points (or ``instances'') during training.  Fig. \ref{fig:BCG_FS_plot} shows all peak locations in a sample BCG signal. In order to form the positive training bags, the 3 instances from each transducer that are closest in time to the ground-truth location marked by the finger sensor were grouped together.  Similarly, all instances that are not placed in a positive bag, are grouped together to form one negative bag. Fig. \ref{fig:pos_bag_plot} plots one randomly picked positive bag, where we can see a positive bag is mixed with false positive data (green plots) and the four assumed true positive data (red plots) are not guaranteed to be all heartbeat signal. By applying the $e$FUMI algorithm to these training bags,  we want to learn a representative and discriminative heartbeat concept. Then, any signature based detection algorithm could be applied for heartbeat detection and heart rate estimation on testing data.

\section{HEARTBEAT CONCEPT LEARNING }\label{sec:3_efumi}

$e$FUMI \cite{Zare:2014whispers, Zare:2015fumi} is a MIL concept learning approach that learns a target concept from imprecisely labeled training data. Specifically, let $\mathbf{X}=\left[\mathbf{x}_1,\cdots,\mathbf{x}_N\right]\in\mathbb{R}^{d\times N}$ be the training data set where $d$ is the dimensionality of an  instance, $\mathbf{x}_i$, and $N$ is the total number of training instances. The data is grouped into $K$ \textit{bags},  $\mathbf{B} = \left\{ B_1, \ldots, B_K\right\}$, with associated binary bag-level labels, $L = \left\{L_1, \ldots, L_K\right\}$ where $L_j \in \left\{ 0, 1\right\}$ and $\mathbf{x}_{i} \in B_j$ denotes the $i^{th}$ instance in bag $B_j$.

Given training data in this form, the goal of $e$FUMI is to estimate a target concept, $\mathbf{e}_T$, non-target concepts, $\mathbf{e}_k, \forall k = 1, \ldots M$ and the representation parameters, $\mathbf{p}_i, \forall ki= 1, \ldots N$, which define the relationship between each data point, $\mathbf{x}_i$, and concepts. Specifically, $e$FUMI treats each data as a convex combination of target and/or non-target concepts, as in \eqref{eq:data_model},

\begin{figure*}[!tb]
	\vspace{+2mm}
	\normalsize
	\begin{equation}
	F = \frac{1}{2}(1-u)\sum_{i=1}^Nw_i\bigg\|(\mathbf{x}_i-z_ip_{iT}\mathbf{e}_T-\sum_{k=1}^Mp_{ik}\mathbf{e}_k)\bigg\|_2^2+\frac{u}{2}\sum_{k=T,1}^{M}\bigg\|\mathbf{e}_k-\boldsymbol{\mu}_0\bigg\|_2^2+\sum_{k=1}^M\gamma_k\sum_{i=1}^Np_{ik}
	\label{eq:datalikelihood}\tag{2}
	\end{equation}
	\begin{small}
		\begin{equation}
		E[F]=\sum_{\substack{z_i\in\{0,1\}}} \left[ \frac{1}{2}(1-u)\sum_{i=1}^N w_i P(z_i|\mathbf{x}_i, \boldsymbol{\theta}^{(t-1)})\left\| \mathbf{x}_i - z_ip_{iT}\mathbf{e}_T - \sum_{k=1}^Mp_{ik}\mathbf{e}_k\right\|_2^2\right]+\frac{u}{2}\sum_{k=T,1}^M\left\|\mathbf{e}_k-\boldsymbol{\mu}_0\right\|_2^2 + \sum_{k=1}^M\gamma_k\sum_{i=1}^Np_{ik}
		\label{eqn:expectation}\tag{3}
		\end{equation}
	\end{small}
	\hrulefill
\end{figure*}

\begin{equation}
\mathbf{x}_i = p_{iT}\mathbf{e}_T + \sum_{k=1}^M p_{ik}\mathbf{e}_k
\label{eq:data_model}
\end{equation}
subject to the non-negative and sum-to-one constraints:

\setcounter{equation}{3}
\begin{equation}
p_{iT} + \sum_{k=1}^M p_{ik} = 1, p_{iT} \ge 0,  p_{ik} \ge 0
\label{eq:convex_constraint}
\end{equation}

Furthermore,  the bag-level labels are \emph{imprecise} since, if for bag $B_j$ with $L_j = 1$ (thus, $B_j$ is a positive bag), this indicates that there is at least one data point in $B_j$ with a non-zero $p_{iT}$, indicating some presence of target concept $\mathbf{e}_T$, as in \eqref{eq:l1}, where $\boldsymbol{\varepsilon}_i$ is a noise term. However, the exact number of data points in a positive bag with a target contribution (i.e., $p_{iT} \ne 0$), is unknown.
\vspace{-2mm} 

\begin{equation}
\text{if }L_j = 1, \exists \mathbf{x}_i \in B_j\text{ s.t. }\mathbf{x}_i = p_{iT}\mathbf{e}_T + \sum_{k=1}^{M} p_{ik}\mathbf{e}_{k}+\boldsymbol{\varepsilon}_{i}, p_{iT} > 0 
\label{eq:l1}
\end{equation}
If $B_j$ is a negative bag (i.e., $L_j = 0$), then this indicates that none of the data in $B_j$ contains any target, as in \eqref{eq:l2}.
\begin{equation}
\text{if }L_j = 0,  \forall \mathbf{x}_i \in B_j, \mathbf{x}_i =  \sum_{k=1}^{M} p_{ik}\mathbf{e}_{k}+\boldsymbol{\varepsilon}_{i}
 \label{eq:l2} \vspace{-2mm}
\end{equation}
Given the model formulation above, the assumed \emph{complete} data log-likelihood is proportional to \eqref{eq:datalikelihood}, where the first term is the approximation error between the training data and the convex combination of estimated concepts and corresponding representations; the second term add a penalty on the distance between estimated concepts and global data mean, to drive the concepts for a tight fit around the data; the third term is a sparsity promoting term used to determine $M$, the number of constituent non-target concepts. In \eqref{eq:datalikelihood}, $z_i$ is the assumed true instance-level labels, $\boldsymbol{\mu}_0$ is the global data mean and $u$ is a regularization parameter; the weight $w_i = \frac{\alpha N^-}{N^+}$ for points in positively labeled bags and  $w_i = 1$ otherwise, where $N^-$ and $N^+$ are the number of points in negative and positive bags, respectively; finally, $\gamma_k = \frac{\Gamma}{\sum_{i=1}^N p_{ik}^{(t-1)}}$ is an adaptive parameter to drive representation values with respect to unnecessary concepts to be pruned to zero  with $\Gamma$ set to a fixed value as in  \cite{Zare:2007}.

To address the fact that there is a set of unknown binary variables $\left\{z_i\right\}_{i=1}^{N^+}$ that account for the true labels for points from positively labeled bags.  $e$FUMI adopts an EM optimization methods by taking the expected value of the log likelihood with respect to $z_i$ as shown in \eqref{eqn:expectation}.  In  \eqref{eqn:expectation}, $p(z_i|\boldsymbol{x}_i, \boldsymbol{\theta}^{(t-1)})$ is estimated given the parameter set estimated in the previous iteration $\boldsymbol{\theta}^{t-1}$ and the constraints of the bag-level labels, $L_j$, as shown in \eqref{eqn:prob}, where $\beta$ is a fixed scaling parameter. Then, the \emph{M-step} is performed by optimizing \eqref{eqn:expectation} for each of the desired parameters.  The algorithm is summarized in Alg. \ref{alg:EM}. In this paper, the parameter settings for $e$FUMI was $u=0.05, M=3, \Gamma=0.1, \alpha=1.5, \beta=5$. Please see \cite{Zare:2015fumi} for more details about physical meaning and setting of $e$FUMI parameters.
\begin{algorithm}
\caption{$e$FUMI EM algorithm}
\algsetup{indent=4em}
\begin{algorithmic}[1] 
\STATE Initialize $\boldsymbol{\theta}^{0} = \left\{ \mathbf{e}_T, \left\{\mathbf{e}_k\right\}_{k=1}^{M}, \left\{\mathbf{p}_i\right\}_{i=1}^{N} \right\}$, $t = 1$
\REPEAT
\STATE \textbf{\emph{E-step}}: Compute  $P(z_i|\mathbf{x}_i, \boldsymbol{\theta}^{(t-1)}), i=1,\cdots,N$ given $\boldsymbol{\theta}^{t-1}$
\STATE \textbf{\emph{M-step}}:
        \STATE Update $\mathbf{e}_T$ and $\left\{\mathbf{e}_k\right\}_{k=1}^{M}$ by minimizing \eqref{eqn:expectation}  wrt.  $\mathbf{e}_T$, $\left\{\mathbf{e}_k\right\}_{k=1}^{M}$
        \STATE  Update $\left\{\mathbf{p}_i\right\}_{i=1}^{N}$ by minimizing \eqref{eqn:expectation} wrt. $\left\{\mathbf{p}_i\right\}_{i=1}^{N}$ s.t. \eqref{eq:convex_constraint}
		\STATE  Prune each $\mathbf{e}_k, k = {1, \ldots, M}$ if $\max_n(p_{nk}) \le \tau$ where $\tau$ is a fixed threshold (e.g. $\tau = 10^{-6}$)\\
\STATE $t \gets t + 1$
\UNTIL{Convergence}
   \RETURN $\mathbf{e}_T, \mathbf{E}, \mathbf{P}$\\
   \footnotesize{*Due to space constraints, the update equations used on steps (5)-(7) and their corresponding derivations are posted here: \url{http://engineers.missouri.edu/zarea/tigersense/code}}
\end{algorithmic} 
\label{alg:EM}
\end{algorithm}
\vspace{-2mm}
\begin{eqnarray}
& &P(z_i|\mathbf{x}_i, \boldsymbol{\theta}^{(l-1)}) = \nonumber \\
& &\left\{ \begin{array}{l l}
e^{-\beta \left\| \mathbf{x}_i - \sum_{k=1}^Mp_{ik}\mathbf{e}_k\right\|_2^2} & \text{if } z_i = 0, L_j = 1\\
1-e^{-\beta \left\| \mathbf{x}_i - \sum_{k=1}^Mp_{ik}\mathbf{e}_k\right\|_2^2} & \text{if } z_i = 1, L_j = 1\\
0 &  \text{if } z_i = 1, L_j = 0\\
1 &  \text{if } z_i = 0, L_j = 0\\
\end{array}\right.
\label{eqn:prob}
\end{eqnarray}

\section{EXPERIMENTAL RESULTS}\label{sec:4_results}

\begin{figure}
\vspace{+2mm}
\begin{center}
\includegraphics[width=6.8cm]{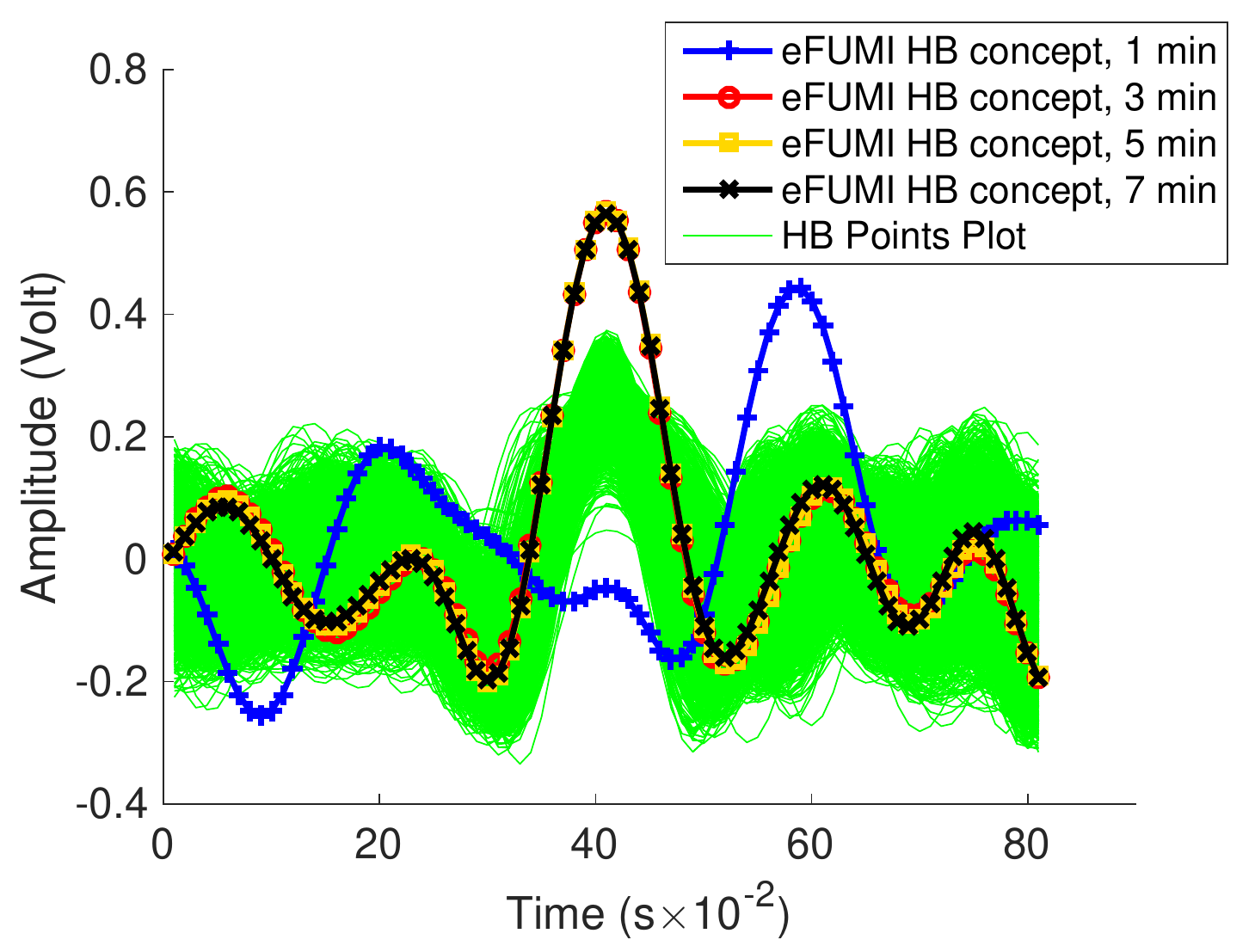}
\caption{Estimated Heartbeat Concept by eFUMI}\label{fig:HB_concepts}
\end{center}
\end{figure}

\begin{figure}
\begin{center}
\includegraphics[width=6.8cm]{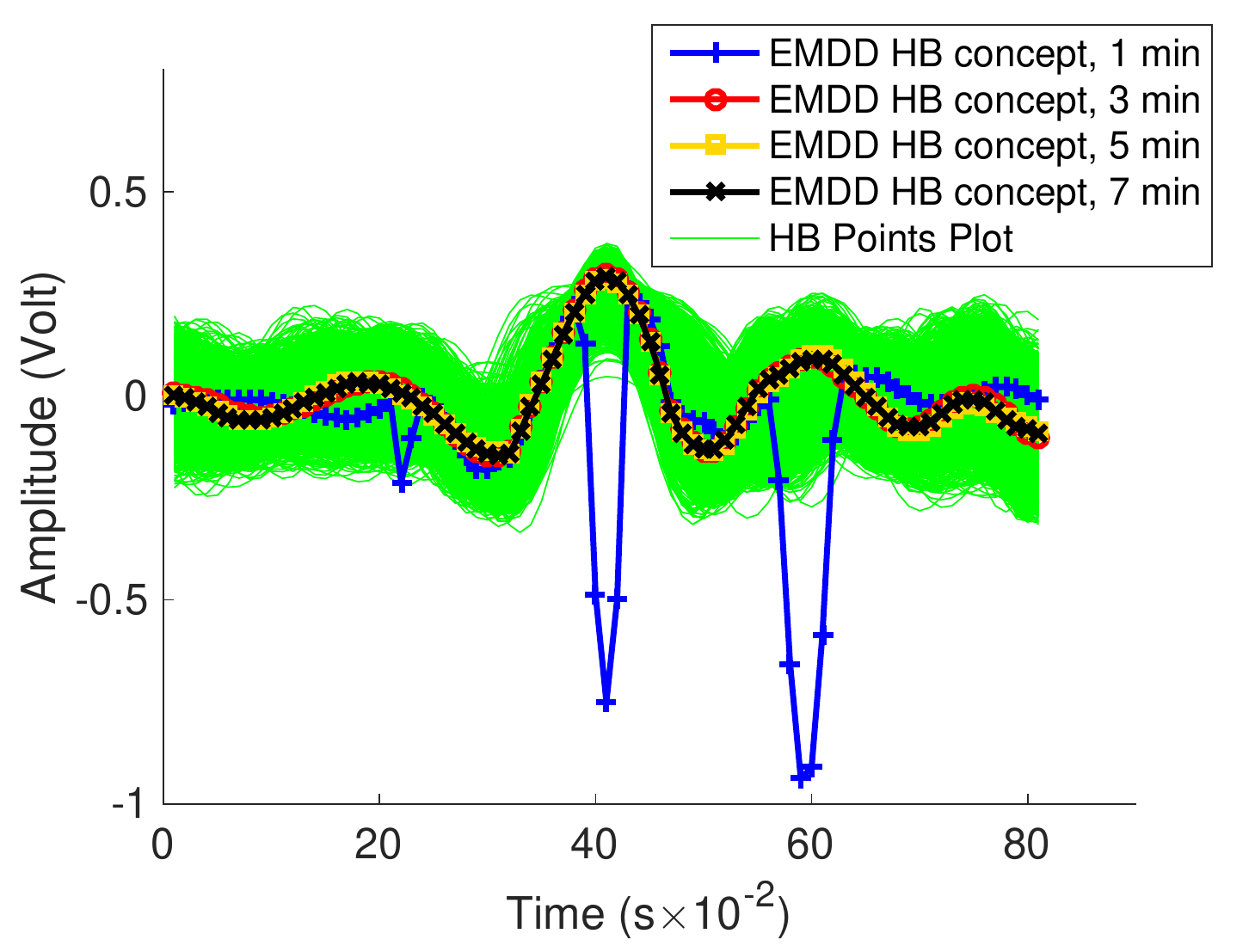}
\caption{Estimated Heartbeat Concept by EMDD}\label{fig:HB_concepts_emdd}
\end{center}
\vspace{-3mm}
\end{figure}

In the first experiment, $e$FUMI was applied to a 10 minute  BCG signal of one randomly selected subject. The goal of this experiment is to determine the amount of training data needed to learn a personalized, discriminative heartbeat concept for an individual.  Specifically, the first 1 minute, 3 minutes, 5 minutes and 7 minutes, respectively, from the beginning of the collected BCG signal were used as training data in this experiment. The last 3 minutes of the BCG signals was held out as test data. The heartbeat concept estimated by $e$FUMI with varying amounts of training data are shown in Fig. \ref{fig:HB_concepts}.  In these results, we can see that except for the 1 minute training set, $e$FUMI is able to learn a heartbeat concept that effectively represents the true heartbeat signals (shown in green). We also applied a comparison MIL algorithm, the Expectation-Maximization version of Diverse Density (EMDD) \cite{Zhang:2002}, to the training data. Fig. \ref{fig:HB_concepts_emdd} shows the heartbeat concepts estimated by EMDD.  In these comparison results, we can see that: (1) the J-peak estimated by EMDD is not as prominent as $e$FUMI; and (2) EMDD also does not work well given only 1 minute of training data.

\begin{figure}
\begin{center}
\includegraphics[width=8.2cm]{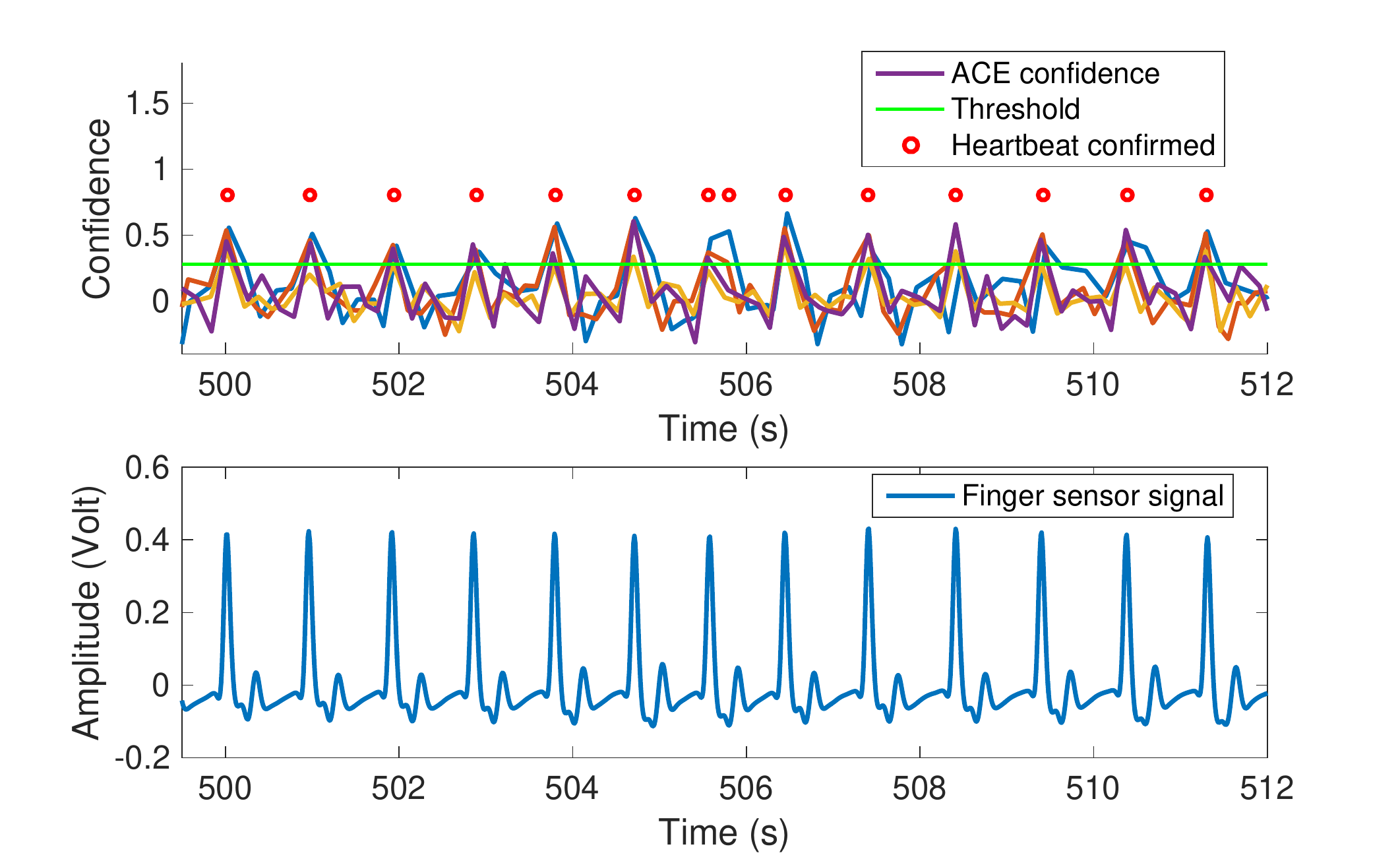}
\caption{Confidence Value and Confirmed Heartbeat}\label{fig:detc_confid}
\end{center}
\vspace{-2mm}
\end{figure}

After learning heartbeat concepts, heartbeat detection on test data can be carried out. In the results shown in this paper, the ACE detector \cite{Kraut:1999, basener:2010clutter} was applied to the test data to get a confidence value for each data point to be a true heartbeat signal. Fig. \ref{fig:detc_confid} shows the plot of the confidence value for test data of four transducers with a concept estimated from 5 minutes of training data.  In this experiment, a heartbeat is confirmed through a voting that if there are at least two confidence values within a neighborhood of $\pm0.03s$ greater than a preset threshold (0.28, determined from training data)  across all four transducers. From Fig. \ref{fig:detc_confid} we can see that the peaks of estimated confidence follow the peaks of the finger sensor signal. Also, a false positive was found around 506s.
 
\begin{figure}
\begin{center}
\subfigure[FPR in Linear]{   
\includegraphics[width=4cm]{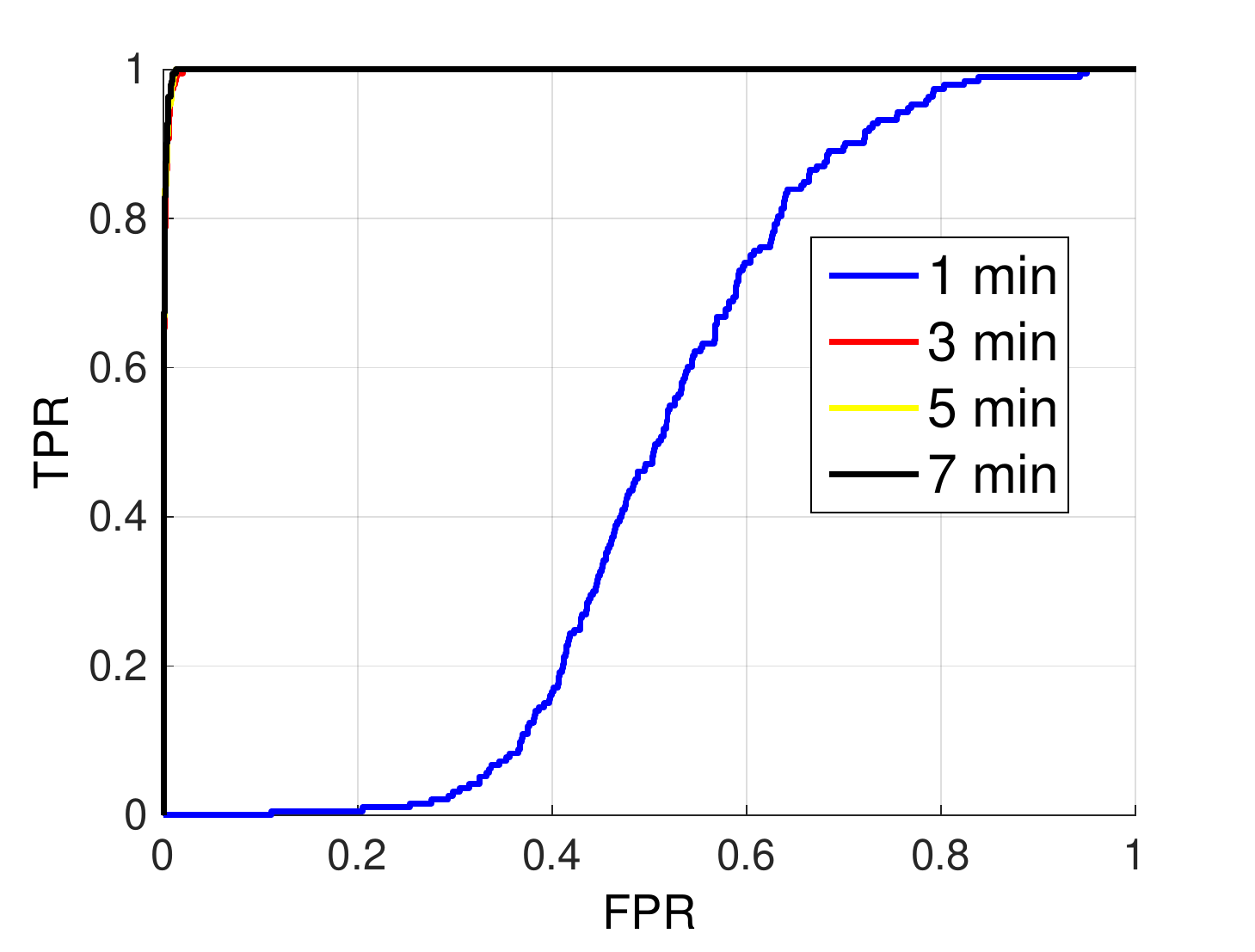} \label{fig:roc_linear}}
\subfigure[FPR in log]{   
\includegraphics[width=4cm]{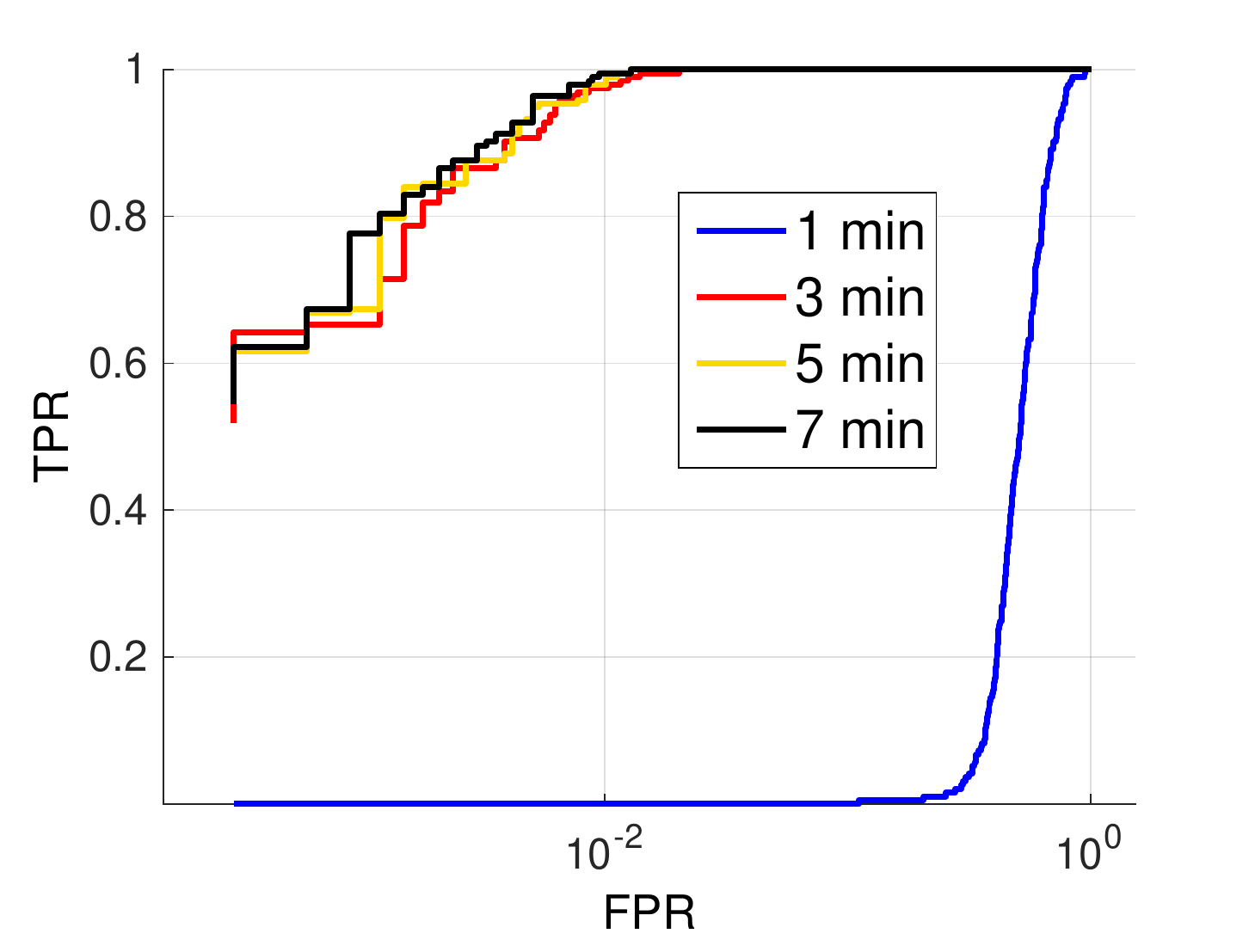} \label{fig:roc_log}}
\caption{Scoring Results of $e$FUMI estimated Heartbeat Concepts}\label{fig:efumi_scoring}
\end{center}
\end{figure}

For quantitative evaluation of heartbeat detection, receiver operating characteristic (ROC) analysis was conducted on confidence estimated by $e$FUMI as in Fig. \ref{fig:efumi_scoring}, where Fig. \ref{fig:roc_linear} shows the True Positive Rate (TPR) vs. False Positive Rate (FPR) in linear scale and Fig. \ref{fig:roc_log} in logarithm scale.

\begin{figure}
\vspace{-5mm}
\begin{center}
% \subfigure[1 Minute Training]{   
% \includegraphics[width=4cm]{HeatRate_est_1min_eFUMI.pdf} \label{fig:rate_1min}}
\subfigure[3 Minute Training]{   
\includegraphics[width=4cm]{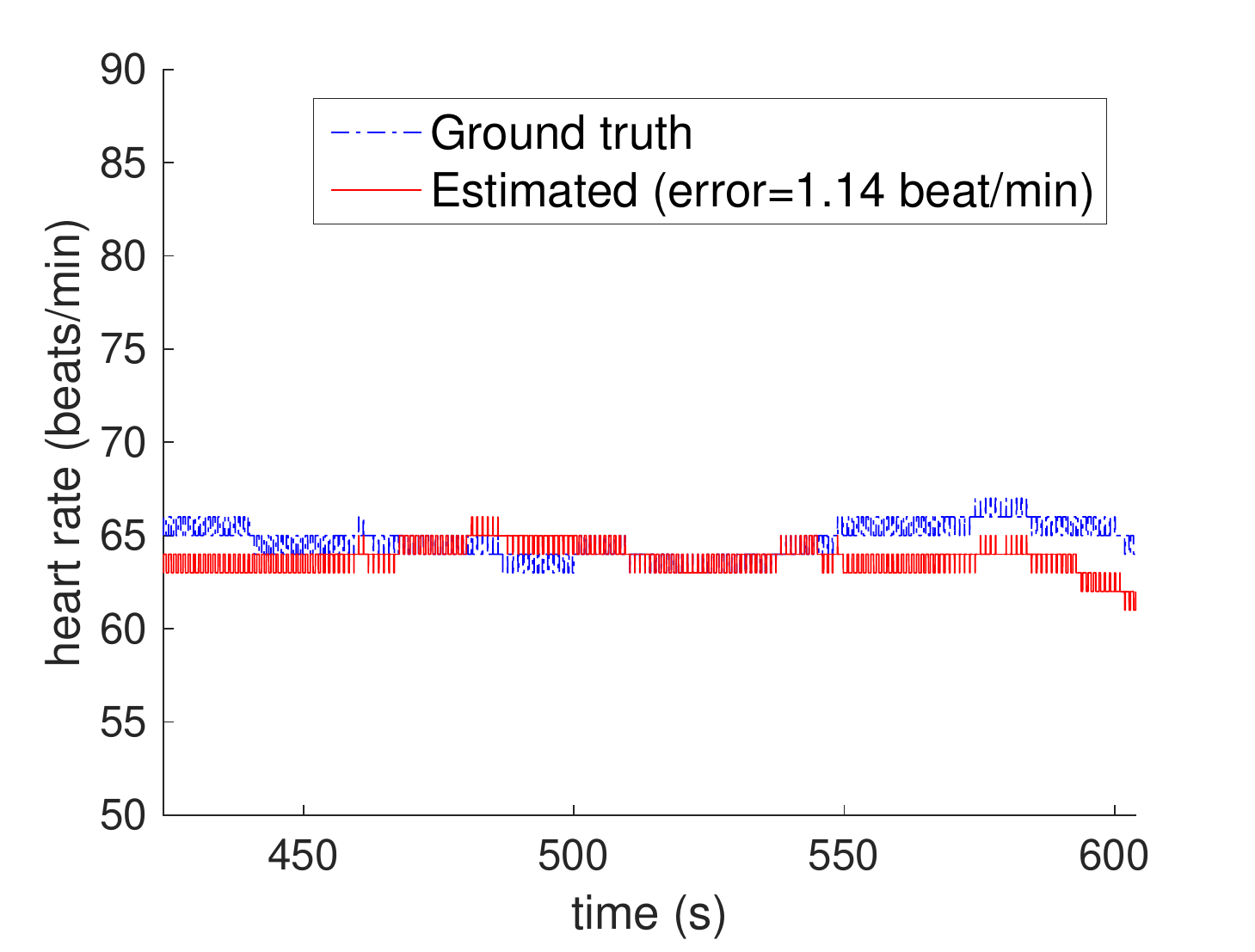} \label{fig:rate_3min}}
\subfigure[5 Minute Training]{   
\includegraphics[width=4cm]{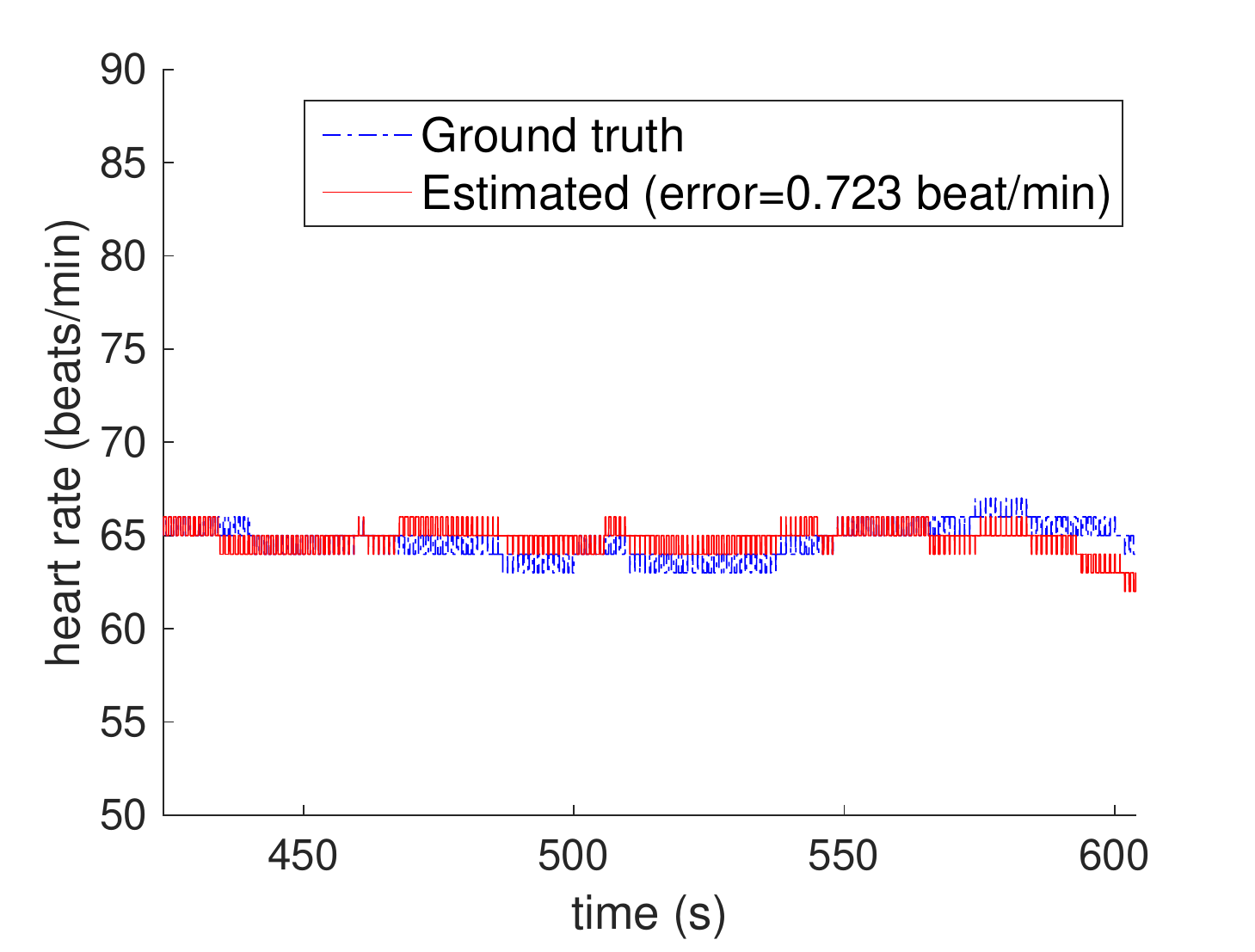} \label{fig:rate_5min}}
% \subfigure[7 Minute Training]{   
% \includegraphics[width=4cm]{HeatRate_est_7min_eFUMI.pdf} \label{fig:rate_7min}}
\caption{Heart Rate Estimation by eFUMI}\label{fig:HB_rate}
\end{center}
\end{figure}
\begin{figure}
\begin{center}
% \subfigure[1 Minute Training]{   
% \includegraphics[width=4cm]{HeatRate_est_1min_EMDD.pdf} \label{fig:rate_1min_emdd}}
\subfigure[3 Minute Training]{   
\includegraphics[width=4cm]{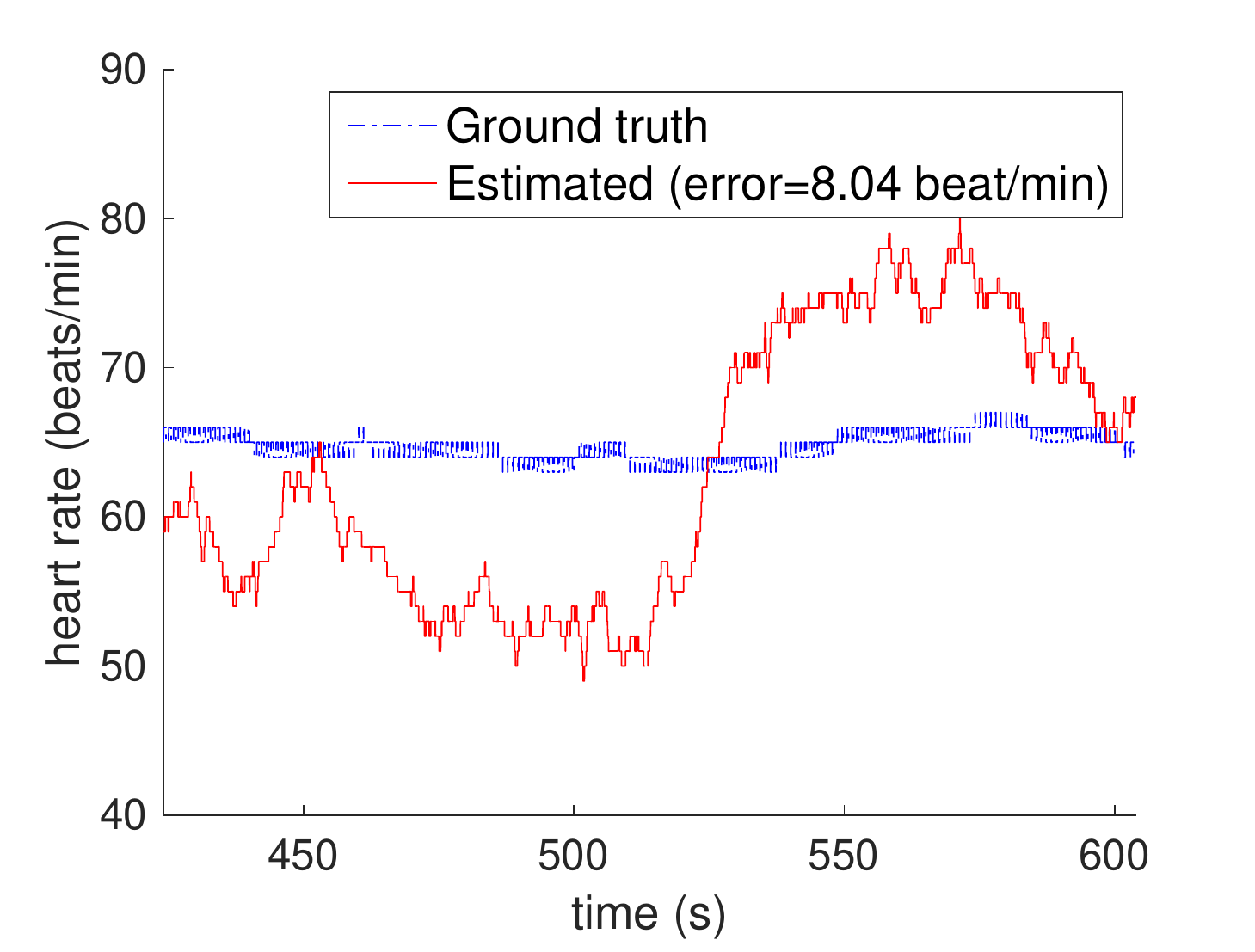} \label{fig:rate_3min_emdd}}
\subfigure[5 Minute Training]{   
\includegraphics[width=4cm]{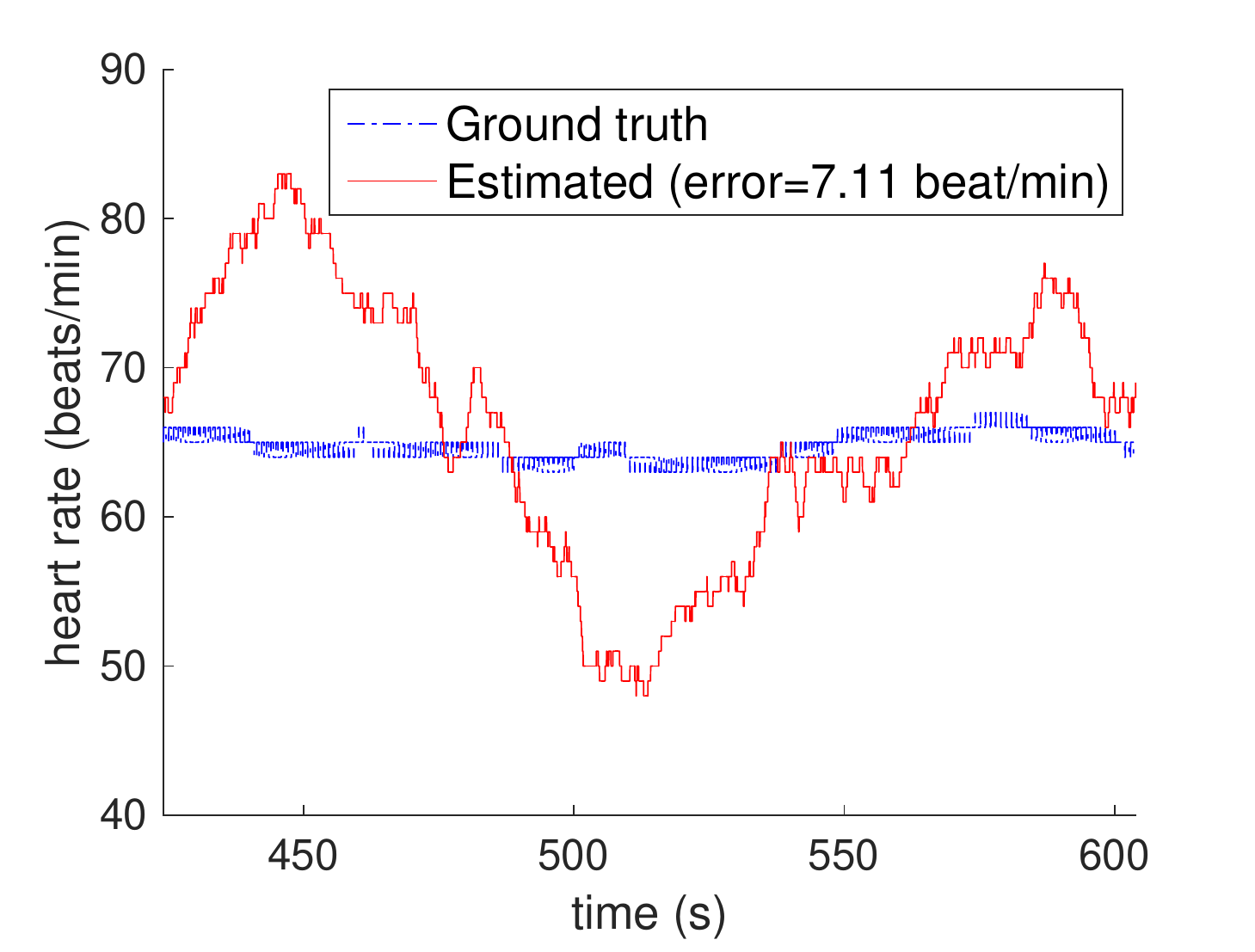} \label{fig:rate_5min_emdd}}
% \subfigure[7 Minute Training]{   
% \includegraphics[width=4cm]{HeatRate_est_7min_EMDD.pdf} \label{fig:rate_7min_emdd}}
\caption{Heart Rate Estimation by EMDD}\label{fig:HB_rate_emdd}
\end{center}
\end{figure}

For heartrate estimation, the average of confirmed heartbeats over 1 minute is computed using a sliding window. Fig. \ref{fig:rate_3min} - Fig. \ref{fig:rate_5min} show the estimated heart rate on testing data by training on length of 3 minutes and 5 minutes data, respectively, denoting the estimation error decreases with the increase in the amount of training. As a comparison, the heart rate estimated by EMDD is shown in Fig. \ref{fig:rate_3min_emdd} - Fig. \ref{fig:rate_5min_emdd}. Table \ref{tab:heart_rate_error} shows the mean error and standard deviation of estimated heart rate.

\begin{figure}
\begin{center}
\includegraphics[width=6.8cm]{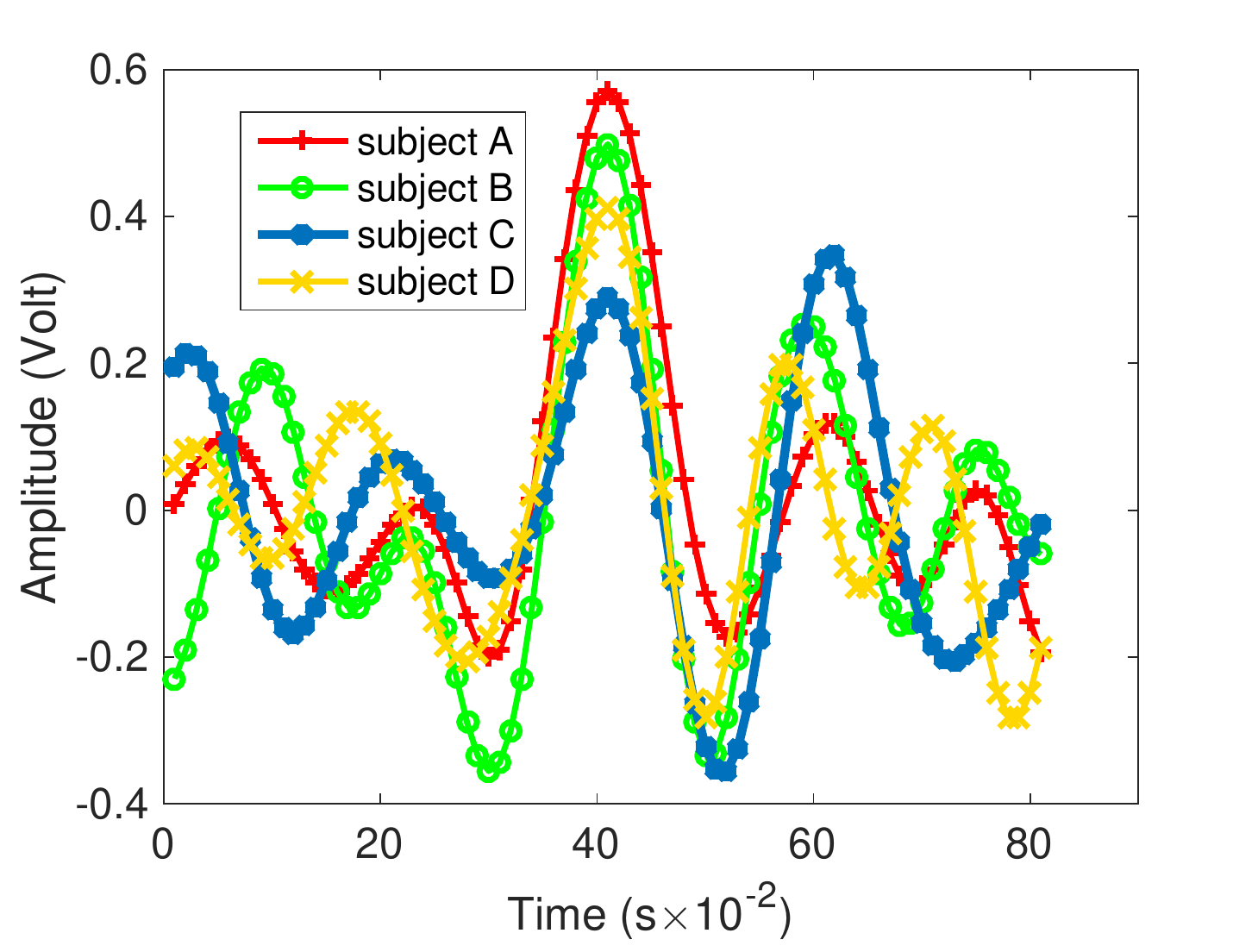}
\caption{Heartbeat Concepts by eFUMI}\label{fig:HB_concepts_5_people_efumi}
\end{center}
\vspace{-3mm}
\end{figure}

\begin{figure}
\vspace{+2mm}
\begin{center}
\includegraphics[width=6.8cm]{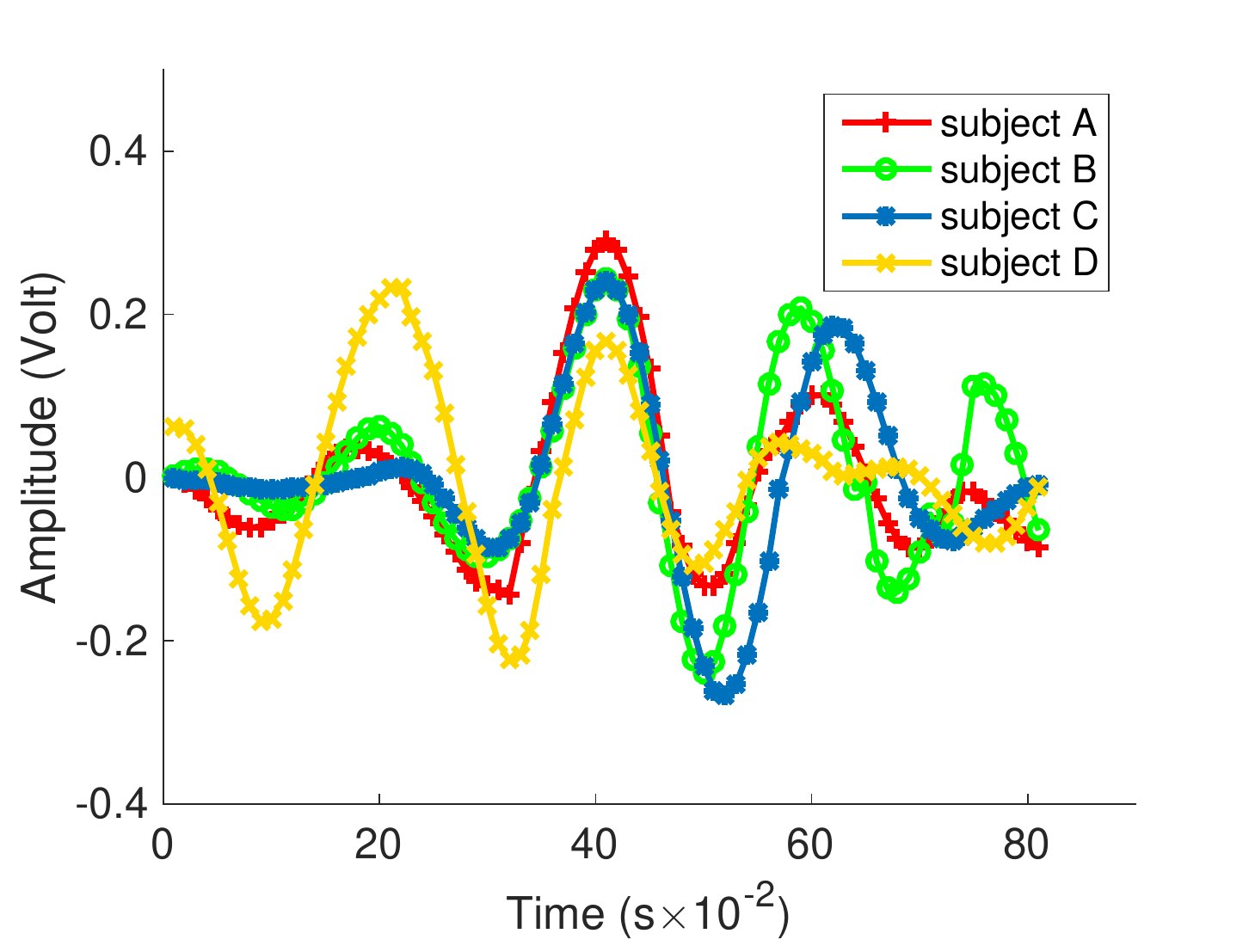}
\caption{Heartbeat Concepts by EMDD}\label{fig:HB_concepts_5_people_emdd}
\end{center}
\vspace{-5mm}
\end{figure}

\begin{table}
\begin{center}
\vspace{5mm}
\caption{Mean error and standard deviation, heart rate estimation}\label{tab:heart_rate_error}
\begin{tabular}{|c|c|c|}
\hline
\multirow{2}{*}{Length of Training} & \multicolumn{2}{c|}{Mean Error (beat/min)} \\
\cline{2-3}
& $e$FUMI & EMDD \cite{Zhang:2002} \\
\hline
\hline
1 minute & $3.31\pm2.86$ & $9.25\pm3.46$ \\
\hline
3 minutes & $1.14\pm0.93$ & $8.04\pm3.45$ \\
\hline
5 minutes & $0.72\pm0.58$ & $7.11\pm4.86$ \\
\hline
7 minutes & $0.63\pm0.50$ & $4.73\pm3.38$ \\
\hline
\end{tabular}
\end{center}
\vspace{-2mm}
\end{table}
In the second experiment, we randomly selected 3 additional subjects.  For each subject, we split the collected BCG signal into 5 minutes for training and 5 minutes for testing. Fig. \ref{fig:HB_concepts_5_people_efumi} shows estimated heartbeat concepts for 4 people by $e$FUMI and Fig. \ref{fig:HB_concepts_5_people_emdd} shows the comparison heartbeat concepts estimated by EMDD on the same data. From Fig. \ref{fig:HB_concepts_5_people_efumi} and Fig. \ref{fig:HB_concepts_5_people_emdd} we can clearly see that the heartbeat concept estimated by $e$FUMI has a more prominent J-peak, which helps improve performance in heartbeat detection and rate estimation. It also shows that there indeed exists variability in heartbeat prototype for different people. Table \ref{tab:heart_rate_error_5_people} shows the mean error and standard deviation of estimated heart rate by both $e$FUMI and EMDD. 
\begin{table}
\begin{center}
\caption{Mean error and standard deviation, heart rate estimation}\label{tab:heart_rate_error_5_people}
\begin{tabular}{|c|c|c|}
\hline
\multirow{2}{*}{Subject No.} & \multicolumn{2}{c|}{Mean Error (beat/min)} \\
\cline{2-3}
& $e$FUMI & EMDD \cite{Zhang:2002} \\
\hline
\hline
subject A & $0.55\pm0.57$ & $6.75\pm4.16$ \\
\hline
subject B & $1.07\pm0.76$ & $4.63\pm2.94$ \\
\hline
subject C & $0.75\pm0.64$ & $4.74\pm3.91$ \\
\hline
subject D & $0.65\pm0.72$ & $5.61\pm3.40$ \\
\hline
\end{tabular}
\end{center}
\vspace{-5mm}
\end{table}

\vspace{-3mm}
\section{CONCLUSION }\label{sec:5_conclusion}
In this paper,  heartbeat characterization was modeled as a MIL problem and addressed using the $e$FUMI algorithm. Experimental results show that with enough  training data, the algorithm is able to perform well in heartbeat characterization and heart rate estimation with errors of less than 1 beat/min. 

\addtolength{\textheight}{-12cm}   % This command serves to balance the column lengths

%%%%%%%%%%%%%%%%%%%%%%%%%%%%%%%%%%%%%%%%%%%%%%%%%%%%%%%%%%%%%%%%%%%%%%%%%%%%%%%%
\vspace{-3mm}
{ \footnotesize \bibliographystyle{IEEEtran}
\bibliography{EMBC_eFUMI_Ref}}

\end{document}